\pdfoutput=1

\documentclass[11pt]{article}

\usepackage[final]{acl}

\usepackage{times}
\usepackage{latexsym}
\usepackage{setspace}  %

\usepackage[T1]{fontenc}

\usepackage[utf8]{inputenc}

\usepackage{microtype}

\usepackage{inconsolata}

\usepackage{graphicx}

\usepackage{microtype}
\usepackage{hyperref}
\usepackage{url}

\usepackage{amsmath,amsfonts,bm}

\def\eqref#1{equation~\ref{#1}}

\def\1{\bm{1}}

\DeclareMathAlphabet{\mathsfit}{\encodingdefault}{\sfdefault}{m}{sl}
\SetMathAlphabet{\mathsfit}{bold}{\encodingdefault}{\sfdefault}{bx}{n}

\usepackage{wrapfig}
\usepackage[framemethod=TikZ]{mdframed}
\usepackage{amsmath}
\usepackage{needspace}
\usepackage{hyperref}
\usepackage{mdframed} %
\usepackage{tabularx} %
\usepackage{makecell} %
\usepackage{booktabs}
\usepackage{siunitx} %
\usepackage{graphicx}  %
\usepackage[font=small,labelfont=bf]{caption}
\usepackage{subcaption}  %
\usepackage{multirow}
\usepackage{tikz}
\usepackage{algorithm}
\usepackage{algpseudocode}
\usepackage{enumitem}

\graphicspath{{figures/}}  %

\usepackage{mathtools} %
\usepackage[font=small,labelfont=bf]{caption}
\usepackage{amsthm}
\theoremstyle{definition}
\newtheorem{definition}{Definition}[section]

\title{CoRAG: Collaborative Retrieval-Augmented Generation}

\author{
  Aashiq Muhamed\textsuperscript{1}, 
  \textbf{Mona Diab\textsuperscript{1}, 
  Virginia Smith\textsuperscript{2}} \\
  \{amuhamed, mdiab, smithv\}@andrew.cmu.edu
  \\
   \textsuperscript{1} Language Technologies Institute,
  \textsuperscript{2} Machine Learning Department \\
  Carnegie Mellon University
  }

\begin{document}
\maketitle

\begin{abstract}

\looseness=-1
Retrieval-Augmented Generation (RAG) models excel in knowledge-intensive tasks, especially under few-shot learning constraints. We introduce CoRAG, a framework extending RAG to collaborative settings, where clients jointly train a shared model using a collaborative passage store. To evaluate CoRAG, we introduce CRAB, a benchmark for collaborative homogeneous open-domain question answering. Our experiments demonstrate that CoRAG consistently outperforms both parametric collaborative learning methods and locally trained RAG models in low-resource scenarios. Further analysis reveals the critical importance of relevant passages within the shared store, the surprising benefits of incorporating irrelevant passages, and the potential for hard negatives to negatively impact performance. This introduces a novel consideration in collaborative RAG: the trade-off between leveraging a collectively enriched knowledge base and the potential risk of incorporating detrimental passages from other clients. Our findings underscore the viability of CoRAG, while also highlighting key design challenges and promising avenues for future research\footnote{Code is available at \url{https://github.com/aashiqmuhamed/CoRAG}}.
\end{abstract}

\section{Introduction}
\looseness=-1
Retrieval-Augmented Generation (RAG) models \citep{lewis2020retrieval, izacard2022fewshot, Qin2019ConversingBR, Zhang2021JointRA}, which incorporate large external datastores of text passages, have shown promise in knowledge-intensive and few-shot tasks. However, their exploration has mainly focused on centralized settings where a single entity controls both the model and the datastore.
 The potential of RAG within a collaborative learning framework, where multiple clients jointly train a shared model without directly exchanging their labeled data \citep{mcmahan2016communicationefficient}, but potentially building a shared passage store, remains largely unexplored. Consider competing businesses in the same industry, each possessing expensive to acquire (labeled) data on customer behavior. Directly sharing these data would be strategically disadvantageous, yet they could collaborate to build a shared passage store of relatively inexpensive (unlabeled) market research documents and economic analyses. This allows them to collectively train a more effective RAG model for market prediction without revealing their valuable labeled data. This approach, particularly in low-resource settings enables them to train a more effective model than any single client could achieve independently.

\looseness=-1
This work introduces CoRAG, a framework for collaborative RAG that enables multiple clients to jointly train a shared model using a collaborative passage store, while allowing them to use their local passage stores during inference.
CoRAG introduces unique challenges stemming from the dynamics of constructing and utilizing this shared store. The composition of this knowledge base, particularly the balance of relevant, irrelevant, and hard-negative passages, significantly impacts the model's performance and generalization capabilities. Our experiments reveal that relevant passages are crucial for model generalization, while hard negatives can be detrimental, and, surprisingly, irrelevant passages can even be beneficial. This introduces a fundamental tension in CoRAG: clients must balance the advantages of a richer, shared knowledge base with the risk of incorporating potentially detrimental passages from others. To explore these dynamics, we introduce CRAB, a homogeneous open-domain question answering benchmark. Using CRAB, we empirically demonstrate that a carefully curated collaborative store, rich in relevant passages and minimizing hard negatives, significantly improves model performance compared to parametric collaborative learning methods and local RAG training.
Our contributions include:
\vspace{-0.1in}
\begin{itemize}[leftmargin=*]
\itemsep0em 
\item \textbf{CoRAG Framework:} We introduce CoRAG, a framework for collaborative training of RAG models. CoRAG enables multiple clients to jointly train a shared model using a collaborative passage store, while allowing the use of client-specific stores during inference. We show that using a collaborative passage store can significantly improve few-shot performance over collaborative parametric or local RAG models. 
\vspace{-0.05in}
\looseness=-1
\item \textbf{Passage Composition and Client Incentives:} We investigate how the composition of the collaborative store (relevant, irrelevant, and hard-negative passages) affects model generalization and client participation incentives. Our analysis uncovers a fundamental tension: clients must weigh the benefits of accessing an enriched collaborative store against the risk of incorporating potentially detrimental passages from other clients.
\end{itemize}

\section{CoRAG Framework}
\label{sec:corag}

\looseness=-1
RAG models  \citep{lewis2020retrieval, izacard2022fewshot} enhance parametric LMs by incorporating external knowledge in the form of a passage store. Given an input x (e.g., a question), a RAG model retrieves relevant documents z from the passage store and uses them to generate an output y (e.g., an answer). The model estimates the probability of generating y given x, denoted as $p_{RAG}(y|x)$, by marginalizing over the top k retrieved documents:
\vspace{-.25in}
\begin{spacing}{1}
{
\small
\begin{align*}
p_{\text{RAG}}(y|x) &\approx \;\;   \sum_{\mathclap{z \in \text{top-}k(R(\cdot|x))}} \;  R(z|x) \prod_{i=1}^{N} G(y_i|z, x, y_{1:i-1}) 
\end{align*}
}
\end{spacing}
\vspace{-.1in}

\looseness=-1
CoRAG (Algorithm \ref{algorithm:corag}) combines collaborative learning with RAG models, enabling clients to jointly train a shared model while leveraging a collaboratively constructed passage store.  This is particularly advantageous in low-resource settings, where individual clients may have limited local data. By pooling their knowledge through a shared passage store, clients gain access to a broader and more diverse knowledge base, facilitating improved learning and generalization.

\mdfsetup{
  backgroundcolor=gray!20,
  roundcorner=5pt,
  innerleftmargin=5pt,
  innerrightmargin=10pt,
  innertopmargin=5pt,
  innerbottommargin=5pt,
  linewidth=0pt %
}
\begin{algorithm}[t!]
\caption{ \small Collaborative Retrieval-Augmented Generation}
\label{algorithm:corag}
\begin{mdframed}
\begin{algorithmic}
\small
\Require $M$ clients, Pretraining data \(D_{\text{pre}}\), Train question answer pairs per client \(\{D_{\text{train},i}\}_{i=1}^M\), Collaborative train passage store \(I_{\text{train}}\), Test passage stores \(\{I_{\text{test},i}\}_{i=1}^M\), Test queries \(\{Q_i\}_{i=1}^M\)
\Ensure Responses \(\{O_i\}_{i=1}^M\)
\State \textit{\textbf{Pretraining:}}
\State Pretrain retriever \(R\) and reader \(G\) using \(D_{\text{pre}}\)
\State \textit{\textbf{Collaborative Training:}}
\For{each round}
    \For{each client \(i\)}
        \State \(R_i, G_i \gets R, G\)  \Comment{Init with global model}
        \State \(P_i \gets R(D_{\text{train},i}, I_{\text{train}})\) \Comment{Retrieve passages}
        \State Update local \(R_i, G_i\) using \(P_i\) and \(D_{\text{train},i}\)
    \EndFor
    \State \(R, G \gets \text{Aggregate}(\{R_i, G_i\}_{i=1}^M)\) \Comment{Update global model}
\EndFor
\State \textit{\textbf{Inference:}}
\For{each client \(i\)}
    \State \(P_i \gets R(Q_i, I_{\text{test},i})\) \Comment{Retrieve client \(i\) passages }
    \State \(O_i \gets G(Q_i, P_i)\) \Comment{Generate client \(i\) response }
\EndFor
\State \Return \(\{O_i\}_{i=1}^M\)
\end{algorithmic}
\end{mdframed}
\end{algorithm}

\looseness=-1
CoRAG operates in three phases: During \emph{Pretraining}, each retriever and reader are pretrained on a large, shared dataset \(D_{pre}\) using self-supervised objectives to enable general language understanding.
In the \emph{Collaborative Learning} phase, clients collaboratively finetune the pretrained retriever and reader on their local training datasets \(\{D_{\text{train},i}\}_{i=1}^M\) by retrieving relevant passages from a collaborative passage store \(I_{train}\), constructed through contributions from all participating clients. 
Client model updates are aggregated in a decentralized or centralized fashion (e.g., using a method such as FedAvg \citep{mcmahan2016communicationefficient}), producing a global model that reflects the collective knowledge gained during collaborative training. 
In the \emph{Inference} phase, clients utilize the collaboratively trained global RAG model to process incoming queries. Each client aims to maximize local question-answering metrics by identifying relevant passages from a local test passage store \(I_{test}\) that may include passages from the collaborative index and new client-specific passages. 

\looseness=-1
In addition to the Reader and Retriever, CoRAG employs the Collaborative Passage Store $I_\text{train}$, a collection of text passages contributed by all participating clients. Separate passage stores are used for training and testing, with their composition (relevant, irrelevant, and hard-negative passages) significantly influencing both model performance and client incentives for contributing high-quality passages, as we will explore further.

\section{Experiments and Results}
\subsection{CRAB: Collaborative RAG Benchmark}
\label{sec:crab}
\looseness=-1
To investigate passage composition in CoRAG, we introduce CRAB, a homogeneous (identically distributed across clients) open-domain QA benchmark derived from NaturalQuestions \citep{kwiatkowski-etal-2019-natural} with train, test, and dev splits distributed across 8 clients. To study few-shot learning, we provide train splits with 16, 32, and 64 sampled training QA pairs per client. 
The unique dev (8752 pairs) and test QA pairs (3600 pairs) are evenly split among clients.

\looseness=-1
The passage datastore for CRAB is derived from the Wikipedia 32M passages (wiki-dec2018) \citep{izacard2022fewshot}. Mirroring real-world scenarios where new documents emerge or shared knowledge becomes inaccessible, CRAB incorporates distinct passage stores for training and testing, ensuring no overlapping passages between them. While test and dev passages are unique to each client, overlaps in relevant passages are possible between different clients. We will release passage stores corresponding to the various passage composition experiments in this work.

\subsection{Experimental Setup}

\looseness=-1
CoRAG is instantiated with Contriever \citep{izacard2021unsupervised} as the retriever and a pretrained T5 base model with Fusion-in-Decoder \citep{izacard2020leveraging} as reader on all 8 clients.  We compare its performance against flan-t5-base \citep{chung2022scaling}, a comparable-sized ($\sim$220M parameters) closed-book (no retrieval) instruction-tuned parametric model.
We focus on smaller models as they are more practical in resource-constrained collaborative learning settings, where communication overhead can be a significant limitation \citep{10.1145/3650203.3663331, nguyen2022begin}.
We pretrained all models on 350 million passages from 2021 Wikipedia and a subset of the 2020 Common Crawl \citep{Thurner2018Scaling}. They are then finetuned  using bloat16 precision using FedAvg on CRAB in few-shot settings (16, 32, and 64 training examples per client). We use the Perplexity Distillation loss \citep{izacard2023atlas} for both pretraining and finetuning. We report the best client-averaged Exact match score (EM) on the test set across rounds, and the micro-averaged metrics for the Centralized baseline. 

We employ the AdamW optimizer with a batch size of 64 and a learning rate of $4 \times 10^{-5}$ with linear decay for both the reader and retriever. The retriever is trained using query-side finetuning. We employ greedy decoding to generate the answers. During both training and testing, we retrieve the top 40 passages and truncate the concatenation of the query and the retrieved passages to a maximum of 384 tokens. 
For \emph{Collaborative Training}, we do not use warmup iterations, train for 10 rounds with 64 epochs per round, and evaluate the model at the end of each round. 
For \emph{Local Training}, we use 20 warmup iterations, train for 1000 steps, and evaluate the model every 100 steps. All models were trained on 4 A6000 GPUs in under a day.
Further details are in Appendix \ref{app:training-details}.

\subsection{CoRAG is Effective in Few-shot Settings}
\begin{figure}[htb]
\centering
\includegraphics[width=\columnwidth,keepaspectratio]{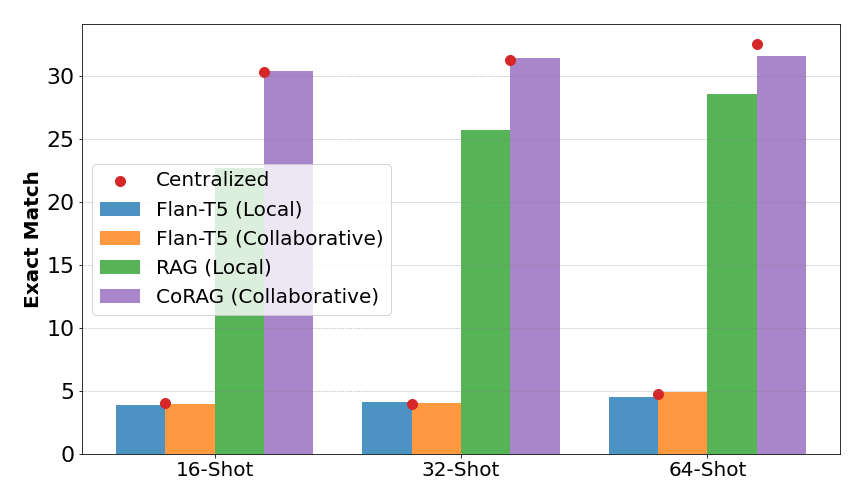}
\caption{Performance of Flan-T5, RAG (Local), and CoRAG on CRAB. CoRAG consistently outperforms Flan-T5 across training configurations. Performance gap between CoRAG and baselines widens as training samples per client decreases.}
\label{fig:few-shot}
\end{figure}

\looseness=-1
Fig \ref{fig:few-shot} compares the few-shot performance of CoRAG against RAG (Local) model and Flan-T5 on CRAB. CoRAG leverages a shared passage store containing the entire Wikipedia, RAG (Local) uses an evenly partitioned Wikipedia across clients to simulate real-world settings, while Flan-T5 relies solely on its parametric knowledge. We evaluate all models in Centralized (combining datasets from all clients), Local (individual client train sets), and Collaborative (locally trained, aggregated after each round) configurations.

\looseness=-1
We find that (i) CoRAG (Collaborative) and RAG (Local) consistently surpass the parametric-only baseline (Flan-T5) in collaborative and local training configurations respectively, across shot settings. (ii) Leveraging the shared passage store confers an advantage to CoRAG over local training. (iii) CoRAG proves particularly effective under limited labeled Q/A pairs per client, showing a 10.5\% improvement over RAG (Local) at 64-shot, which increases to 33.8\% at 16-shot. (iv) CoRAG performance is close to Centralized, consistent with previous observations in benchmarks with homogeneous (identically distributed) client data. These results establish CoRAG as a promising direction for few-shot learning.

\subsection{Impact of Passage Store Composition}
\label{sec:passagecompmain}
\begin{figure*}[htb]
    \centering
    \includegraphics[width=\textwidth,height=0.2\textheight,keepaspectratio]{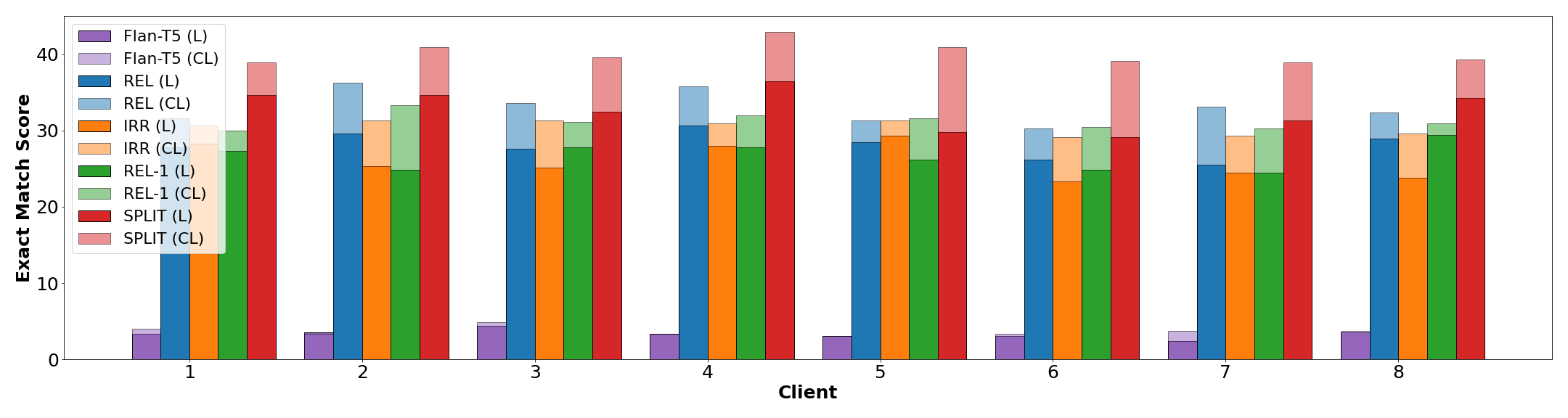}
    \caption{64-shot EM scores on the CRAB benchmark. L is Local and CL is Collaborative. CoRAG consistently improves over RAG (Local) across all clients (1-8) and store choices. Improvement varies depending on the composition of passage store.}
    \label{fig:incentive}
\end{figure*}

\looseness=-1
We investigate how the \emph{train} passage store composition impacts few-shot QA performance.
We classify the BM25-retrieved passages for each concatenated QA pair as a query. The passages are categorized as relevant (top-5 passages containing the ground truth answer), hard negatives (ranked 6–50), and irrelevant (all remaining passages). To validate our categorization, we manually inspected 100 question-answer pairs and confirmed that our chosen ranges effectively captured the intended distinctions.
We construct four train passage stores: (1) REL: Collaborative store containing relevant passages for all client QA data + 80\% of Wikipedia (2) IRR: Collaborative store containing 80\% of Wikipedia, but excluding all relevant passages (3) REL-1: Seven clients use IRR; one  client uses IRR + relevant passages for all client QA data (4) SPLIT: Each client store has relevant passages for their own QA data + 10\% of Wikipedia. The disjoint test sets $I_\text{test}$  are client-local and comprise relevant passages for the test QA data and 2.5\% of Wikipedia.

\looseness=-1
Table \ref{tab:index-perf} compares the 64-shot performance of RAG (Local) and CoRAG on the four store variants. CoRAG consistently outperforms RAG (Local) across all train store variants, and matches the Centralized RAG baseline. The presence of relevant passages in REL significantly improves performance over IRR, confirming their importance for generalization. Interestingly, concentrating relevant passages in a single client (REL-1) only marginally improves over IRR. This is because the benefits manifest through indirect information flow: relevant passages improve client 8's generalization (see Figure \ref{fig:incentive}), which then propagates to other clients via collaborative training. Finally, SPLIT, with a higher concentration of client-specific relevant passages, further boosts performance, highlighting the benefits of selectively concentrating relevant passages during training.

\begin{table}[tb]
\centering
\small
\begin{tabular}{@{}lcccc@{}}
\toprule
\textbf{Passage Store $\rightarrow$} & REL & IRR & REL-1 & SPLIT \\
\midrule
RAG (Local) & 28.088 & 25.944 & 26.597 & 34.694 \\
CoRAG & 33.011 & 30.444 & 30.944 & 40.056 \\
\bottomrule
\end{tabular}
\caption{ Average EM under various passage store options. CoRAG outperforms RAG (Local). REL outperforms IRR, highlighting the importance of relevant passages. SPLIT outperforms REL, showing the benefit of passage concentration.}
\label{tab:index-perf}
\end{table}

\looseness=-1
Table \ref{tab:passage-comp-L} analyzes how training passage store composition affects RAG (Local) performance. Randomly downsampling irrelevant and hard-negative passages from REL has minimal impact. Notably, including hard negatives during training generally decreases performance, while irrelevant passages tend to improve performance. 

\looseness=-1
Our initial investigation suggests two possible mechanisms underlying these trends. First, from the retriever’s perspective, hard negatives introduce ambiguity in non-contrastive RAG training, as their partial lexical and semantic overlap with gold passages generates weak or contradictory gradient signals. Unlike contrastively trained retrievers, which explicitly optimize for hard negative separation, the end-to-end RAG training framework lacks a structured push-away mechanism, leading to suboptimal passage ranking. In contrast, irrelevant passages act as easy negatives, creating a cleaner decision boundary between relevant and non-relevant documents, thereby reinforcing retriever robustness. Second, from the reader’s perspective, irrelevant passages may mitigate entropy collapse, a failure mode in which excessively low attention entropy causes the model to overcommit to misleading context. This more diffuse distribution of attention ultimately improves test-time RAG performance \citep{cuconasu2024power}.

\begin{table}[htb]
\centering
\small
\begin{tabular}{@{} lcc@ {}}
\toprule
\textbf{Train Passage Store Composition} & \textbf{Exact Match} \\
\midrule
Only relevant & 29.111 \\
Only hard neg + irrelevant & 25.222  \\
Only relevant + hard neg & 25.778  \\
Only relevant + irrelevant & \textbf{32.667} \\
Only top-1 relevant + irrelevant & 31.556 \\
\bottomrule
\end{tabular}
\caption{ \looseness=-1 Effect of training passage store composition on RAG (local) test performance averaged across 8 clients. Hard negatives hurt performance, while irrelevant passages are surprisingly beneficial.}
\label{tab:passage-comp-L}
\end{table}

\subsection{Client Incentives}
\looseness=-1
We observe in \autoref{fig:incentive} that CoRAG outperforms RAG (Local) across all passage stores, with gains varying based on store composition. This introduces a novel challenge in CoRAG: strategically deciding which passages to contribute. Unlike traditional collaborative learning, CoRAG introduces a tension between maximizing individual utility and contributing to the collective knowledge base. Contributing high-quality passages benefits all clients but risks incorporating detrimental hard negatives from others. Clients with many relevant passages might be reluctant to contribute, fearing dilution of their advantage, while those with fewer relevant passages stand to gain more from collaboration.

\looseness=-1
The decision to contribute balances potential improvements from accessing a larger passage pool against the risk of incorporating hard negatives. Appendix \ref{app:formal-incentives} formalizes this trade-off in a client utility model. Addressing this tension requires designing mechanisms that incentivize high-quality contributions while ensuring equitable participation, such as contribution-based rewards, tiered access levels, and reputation systems to track client contribution history.

\section{Conclusion and Future Work}
\looseness=-1
This work introduces CoRAG, a framework extending RAG to collaborative learning, enabling clients to jointly train a shared model and collaboratively construct a passage store. Our experiments on CRAB, a collaborative QA benchmark, demonstrate the significant performance advantage of CoRAG in few-shot settings. We analyze the impact of passage store composition on performance, highlighting the importance of relevant and, surprisingly, irrelevant passages, while showing the detrimental effects of hard negatives. Future work includes evaluating CoRAG on heterogeneous client distributions, and designing robust incentive mechanisms.

\section*{Acknowledgements}
This work was supported in part by the National Science Foundation grants IIS2145670 and CCF2107024, and funding from Amazon, Apple, Google, Intel, Meta, and the CyLab Security and Privacy Institute. Any opinions, findings and conclusions or recommendations expressed in this material are those of the author(s) and do not necessarily reflect the views of any of these funding agencies.

\section{Limitations}

Our work presents a promising step towards collaborative RAG, but it is important to acknowledge its limitations and highlight areas for future research. 

\paragraph{Homogeneous Data Distribution.} Our experiments focus on a homogeneous setting where clients have identically distributed data. This simplification allows us to isolate the impact of passage composition and client incentives. However, real-world collaborative scenarios often involve heterogeneous data distributions, where clients possess data from different sources, domains, or with varying levels of quality.  Evaluating CoRAG's effectiveness and fairness under heterogeneous settings is am important area for future work.

\paragraph{Scalability and Efficiency.}  Our experiments are conducted on a relatively small scale with 8 clients. Scaling CoRAG to a larger number of clients, potentially with diverse computational resources and communication constraints, presents challenges related to communication efficiency, model aggregation, and handling of large passage stores.  Exploring optimization strategies to enhance scalability is a promising direction for future research.

\paragraph{Incentive Mechanism Design.}  We propose potential incentive mechanisms to address the tension between individual utility and contributing to the common good. However, designing, evaluating, and deploying robust incentive mechanisms that effectively promote high-quality contributions while ensuring fairness requires further investigation. 

\section{Ethical Considerations}

\looseness=-1
While CoRAG offers promising benefits for few-shot collaborative model training, we acknowledge and address the potential ethical considerations associated with its development and deployment. 

\paragraph{Bias.} The shared passage store, constructed collaboratively by multiple clients, may inadvertently reflect biases present in the data held by individual clients. This could lead to unfair or discriminatory outcomes, particularly if the trained model is used in applications that impact decision-making. Mitigating this risk requires developing robust mechanisms for bias detection and mitigation during the construction and maintenance of the shared store.

\looseness=-1
\paragraph{Misuse.} The capabilities of CoRAG could be exploited for malicious purposes, such as generating harmful or misleading content. Safeguards against such misuse are essential and could include access control mechanisms, content moderation strategies, and clear ethical guidelines for using the technology.  

\paragraph{Equity and Fairness.} The benefits of collaborative RAG should be accessible to all participating clients, regardless of their data resources or technical capabilities. This requires designing incentive mechanisms that encourage contributions from a diverse range of clients and providing support to those with limited data or expertise to ensure equitable participation. 

\looseness=-1
Addressing these ethical considerations throughout the design, development, and deployment of CoRAG systems can help ensure their responsible use.

\subsection*{Data \& Licensing Considerations}

To ensure reproducibility and facilitate further research in collaborative retrieval-augmented generation, we release the following resources under permissive licenses:

\begin{itemize} [noitemsep, leftmargin=10pt, topsep=2pt, partopsep=2pt]
\item \textbf{CoRAG Codebase:} The complete codebase for implementing CoRAG, including the retriever, reader, training procedures, and code for generating the different passage store variants. 
\looseness=-1
\item \textbf{CRAB Dataset:} The CRAB benchmark dataset, including the data splits, the passage datastore, and the evaluation scripts. This dataset is constructed using the NaturalQuestions dataset, which is released under the Apache License 2.0, and the Wikipedia 32M passages (wiki-dec2018) dataset, which is publicly available. Our use of these datasets is consistent with their intended use and licensing terms.  
 
\end{itemize}
\looseness=-1
We have documented configurations, prompt details, training procedures, and hyperparameter selection in \autoref{app:training-details}, to ensure reproducibility. All publicly available datasets used in this work have followed accepted privacy practices at the time of their creation.

\bibliography{custom}

\appendix

\clearpage 
\section{Related Work}
\label{app:related-work}

\paragraph{Collaborative Learning.} Collaborative learning (CL) \citep{mcmahan2016communicationefficient, cho2022maximizing, huang2023evaluating, haghtalab2022demand, karimireddy2022mechanisms} enables multiple clients to jointly train a shared model without directly sharing their raw data.  Traditional CL methods primarily focus on parametric models, where the shared model is represented by a set of parameters that are updated iteratively based on client contributions. 

\paragraph{Retrieval-Augmented Generation.}  RAG models \citep{lewis2020retrieval, izacard2022fewshot, gao2023retrieval} augment parametric language models with a large external datastore of text passages, enabling them to access and utilize a richer knowledge base.  Centralized RAG has shown impressive performance in various tasks, including few-shot learning, open-ended question answering, and knowledge-grounded generation. 

\paragraph{Data-Centric RAG.} Recent works have explored the impact of context composition on RAG performance at inference time \citep{cuconasu2024power, pickett2024better, fatehkia2024trag, he2024position}. For example, \citet{cuconasu2024power} demonstrated that incorporating irrelevant passages during inference can improve generalization.  Our work investigates this phenomenon during \emph{training} within a collaborative setting, studying the role of passage composition.

\paragraph{Privacy-Preserving RAG.}  Recent work has explored using RAG to enhance privacy and compliance in centralized settings. \citet{min2023silo} proposed Silo-LM, a language model that trains a parametric component on low-risk data and uses a separate nonparametric datastore for high-risk data, only accessing the latter during inference.  \citet{wutschitz2023rethinking} investigated privacy in language modeling from an information flow control perspective, finding that RAG offers superior utility and scalability while maintaining perfect secrecy.

\noindent Our work builds upon existing work by:
\begin{itemize}  [noitemsep, leftmargin=10pt, topsep=2pt, partopsep=2pt]
    \item Introducing CoRAG, a novel framework for collaborative RAG that enables clients to jointly train a shared model and leverage a collaboratively constructed passage store.
    \item Systematically analyzing the data-centric aspects of collaborative RAG, focusing on the impact of passage composition on both model generalization and client incentives. 
    \item Highlighting the unique challenges related to passage contribution in collaborative RAG and proposing potential directions for incentive mechanism design to address these challenges. 
\end{itemize}

\section{Training Details and Hyperparameters}
\label{app:training-details}

For question answering on the CRAB benchmark, we format the input using the following template:

\begin{verbatim}
question: {question text} answer: [MASK_0]
\end{verbatim}

\noindent The model is then trained to generate the masked token followed by the answer:

\begin{verbatim}
[MASK_0] {answer}.
\end{verbatim}

We employ greedy decoding to generate the answers. During both training and testing, we retrieve the top 40 passages and truncate the concatenation of the query and the retrieved passages to a maximum of 384 tokens.  

\paragraph{Hyperparameter Settings.} All models are trained using bfloat16 precision. For both the parametric baseline (Flan-T5-base) and CoRAG, we employ the AdamW optimizer with a batch size of 64 and a learning rate of $4 \times 10^{-5}$ with linear decay for both the language model and the retriever. The retriever is trained using query-side fine-tuning. 

\paragraph{Training Procedures.} The training procedures for collaborative and local settings differ slightly. Unless otherwise specified, we report the average of three runs.

\noindent \emph{Collaborative Training:}  We do not use warmup iterations, train for 10 rounds with 64 epochs per round, and evaluate the model at the end of each round. 
For collaborative training, we utilize FedAvg \citep{mcmahan2016communicationefficient} for model aggregation at the server, and we train on 8 clients. 

 \noindent \emph{Local Training:} We use 20 warmup iterations, train for 1000 steps, and evaluate the model every 100 steps.

\paragraph{Compute} All models were trained on 4 A6000 GPUs in under a day. We use exact MIPS search using FAISS \citep{douze2024faiss}, and all indices can be constructed in under 8 hours on a single A6000.

\section{Pretraining Data}
\looseness=-1
Both CoRAG and RAG (Local) retriever and reader are pretrained on a datastore consisting of 350 million passages from the 2021 Wikipedia dump and a subset of the 2020 Common Crawl dump \citep{Thurner2018Scaling}.  This pretraining aims to provide a strong foundation for general language understanding.

The parametric Flan-T5-base model used in our experiments was also pretrained on Common Crawl \citep{wenzek2019ccnet}, which includes English Wikipedia. While this pretraining provides general language capabilities, these models generally do not perform well on open-domain question-answering benchmarks like NaturalQuestions without further fine-tuning. This is because the pretraining data and objectives are not specifically tailored for open-domain question answering.

\section{Few-Shot Performance on CRAB}
\label{app:few-shot}

Table \ref{tab:appendix_model_performance} reports the performance of Flan-T5, T5-base, and RAG (Local and Collaborative) on the CRAB benchmark in few-shot settings. 

\begin{table*}[htb]
\small
\centering
\begin{tabular}{
  l
  S[table-format=2.3]
  S[table-format=2.3]
  S[table-format=2.3]
  S[table-format=2.3]
  S[table-format=2.3]
  S[table-format=2.3]
}
\toprule
& \multicolumn{2}{c}{T5-base} & \multicolumn{2}{c}{Flan-T5-base} & \multicolumn{2}{c}{RAG} \\
\cmidrule(lr){2-3} \cmidrule(lr){4-5} \cmidrule(lr){6-7}
& {EM $\uparrow$} & {F1 $\uparrow$} & {EM $\uparrow$} & {F1 $\uparrow$} & {EM $\uparrow$} & {F1 $\uparrow$} \\
\midrule
Centralized (64-shot) & 3.340 & 6.892 & 4.810 & 8.678 & 32.556 & 41.071 \\
Local (64-shot)       & 3.084 & 6.531 & 4.584 & 8.350 & 28.639 & 36.178 \\
Collaborative (64-shot)   & 3.627 & 7.199 & 4.944 & 8.770 & 31.639 & 39.900 \\
\addlinespace
Centralized (32-shot) & 2.880 & 6.292 & 4.011 & 7.933 & 31.324 & 39.250 \\
Local (32-shot)       & 2.572 & 5.938 & 4.138 & 8.175 & 25.722 & 33.630 \\
Collaborative (32-shot)   & 2.910 & 6.410 & 4.038 & 8.010 & 31.472 & 39.439 \\
\addlinespace
Centralized (16-shot) & 2.810 & 5.810 & 4.033 & 7.650 & 30.320 & 38.164 \\
Local (16-shot)       & 2.610 & 5.456 & 3.916 & 7.388 & 22.722 & 30.256 \\
Collaborative (16-shot)   & 2.890 & 6.099 & 4.021 & 7.820 & 30.416 & 38.218 \\
\bottomrule
\end{tabular}
\caption{Few-shot test performance of RAG and parametric models (T5-base and Flan-T5-base) on the CRAB benchmark across different training strategies and shot levels. CoRAG (RAG Collaborative) consistently outperforms parametric models. Collaborative training yields more substantial improvements for RAG than for parametric models, with the performance gap widening as the number of training samples decreases.}
\label{tab:appendix_model_performance}
\end{table*}

\noindent Table \ref{tab:appendix_model_performance_dev} presents the corresponding performance on the CRAB development set.

\begin{table*}[htb]
\centering
\small
\begin{tabular}{
  l
  S[table-format=2.3]
  S[table-format=2.3]
  S[table-format=2.3]
  S[table-format=2.3]
  S[table-format=2.3]
  S[table-format=2.3]
}
\toprule
Model name & \multicolumn{2}{c}{Centralized} & \multicolumn{2}{c}{Local} & \multicolumn{2}{c}{Collaborative} \\
\cmidrule(lr){2-3} \cmidrule(lr){4-5} \cmidrule(lr){6-7}
 & {Exact Match $\uparrow$} & {F1 $\uparrow$} & {Exact Match $\uparrow$} & {F1 $\uparrow$} & {Exact Match $\uparrow$} & {F1 $\uparrow$} \\
\midrule
T5-base       & 1.862 & 4.986 & 1.302 & 3.814 & 2.057 & 5.343 \\
Flan-T5-base  & 3.142 & 7.069 & 2.959 & 6.852 & 3.736 & 7.956 \\
RAG    & 32.735 & 41.594 & 28.222 & 37.219 & 31.936 & 41.125 \\
\bottomrule
\end{tabular}
\caption{Few-shot performance of parametric models and RAG on the CRAB development set. CoRAG (RAG Collaborative) consistently outperforms the parametric models.}
\label{tab:appendix_model_performance_dev}
\end{table*}

\section{Impact of Passage Store Composition}
\label{app:passage-comp}

To better understand the impact of passage store composition on local RAG performance, we evaluated the client model's performance after adjusting the composition of the REL passage store \(I_\text{train}\) in Table \ref{tab:appendix-index-composition}. Recall that the REL store contains all relevant passages for the training data. In addition to the results in \autoref{sec:passagecompmain}, this table presents results where the relevant passages are kept constant, while the irrelevant and hard-negative passages are uniformly subsampled. This subsampling, which maintains the original proportion of hard negatives to irrelevant passages, has minimal impact on performance. We also observe that removing relevant passages during training is less detrimental than removing them during inference, as the test passage store always contains relevant passages.

Our analysis reveals a nuanced impact of passage store composition on local RAG performance. Incorporating hard negatives into the collaborative store generally leads to lower Exact Match and F1 scores. This suggests that hard negatives, despite their similarity to relevant passages, can mislead the retriever during training, leading to reduced performance at inference time. This differs from the findings in the contrastive learning literature, where hard negatives can be beneficial. In general, the composition of collaborative passages during training can affect test-time performance in several ways: (1) Distribution Shift: there is a shift between the collaborative passage store used during training and the client-specific passage stores used at inference. (2) Retriever Generalization: improving the training composition can enhance the retriever's ability to identify relevant passages at test time. (3) Reader Utilization: a better training composition can also improve the reader's ability to utilize those retrieved passages effectively. However, as CoRAG fine-tuning is not contrastive, it treats all retrieved passages equally, leading to reduced performance when hard negatives similar to relevant passages are present during training. However, including irrelevant passages in the collaborative store that are easier to distinguish often improves performance, indicating their potential role in helping the retriever learn to discriminate between relevant and irrelevant information.

\begin{table*}[htb]
\centering
\small
\begin{tabular}{@{}lcc|cc@{}}
\toprule
\textbf{Passage Store Composition} & \multicolumn{2}{c|}{\textbf{Test Store Only}} & \multicolumn{2}{c}{\textbf{Test+Train Store}} \\
\cmidrule(lr){2-3} \cmidrule(l){4-5}
 & \textbf{Exact Match $\uparrow$} & \textbf{F1 $\uparrow$} & \textbf{Exact Match $\uparrow$} & \textbf{F1 $\uparrow$} \\
\midrule
100\% store & 31.111 & 39.760 & 29.333 & 37.249 \\
80\% store (relevant + others) & 30.222 & 38.685 & 28.667 & 35.525 \\
50\% store (relevant + others) & 31.111 & 39.015 & 29.333 & 37.034 \\
20\% store (relevant + others) & 31.778 & 40.835 & 28.444 & 35.647 \\
10\% store (relevant + others) & 31.111 & 38.969 & 30.222 & 37.503 \\
1\% store (relevant + others)  & 29.333 & 37.418 & 30.889 & 39.233 \\
0\% store & 23.778 & 29.689 & 20.889 & 26.712 \\
Only relevant & 29.111 & 36.467 & 28.667 & 38.597 \\
Only hard neg + irrelevant & 25.222 & 32.046 & 25.556 & 32.063 \\
Only relevant + hard neg & 25.778 & 32.093 & 27.111 & 33.441 \\
Only relevant + irrelevant & 32.667 & 40.569 & 30.111 & 36.969 \\
Only top-1 relevant + irrelevant & 31.556 & 40.890 & 30.333 & 37.703 \\
\bottomrule
\end{tabular}

\caption{Performance comparison of RAG (local) across various training store compositions. We assess the impact on Exact Match and F1 scores at test time, using the local test store ($I_\text{test}$) only and the combined test and train stores ($I_\text{test}$ + $I_\text{train}$  ). Scores are averaged across 8 clients.}
\label{tab:appendix-index-composition}
\end{table*}

\section{Client-Specific Performance Gains on CRAB}
\label{app:client-incentives}

Table \ref{tab:appendix-client-incentives} presents the per-client performance gain of CoRAG over RAG (Local) for the various passage store configurations in the CRAB benchmark. This data was used to generate Figure \ref{fig:incentive}, which visually depicts the impact of collaboration on individual client performance.

\begin{table*}[htb]
\centering
\resizebox{\textwidth}{!}{%
\begin{tabular}{@{}lcccccccccccccccc@{}}
\toprule
\textbf{Passage Store} & \multicolumn{2}{c}{Client 1} & \multicolumn{2}{c}{Client 2} & \multicolumn{2}{c}{Client 3} & \multicolumn{2}{c}{Client 4} & \multicolumn{2}{c}{Client 5} & \multicolumn{2}{c}{Client 6} & \multicolumn{2}{c}{Client 7} & \multicolumn{2}{c}{Client 8} \\
\cmidrule(lr){2-3} \cmidrule(lr){4-5} \cmidrule(lr){6-7} \cmidrule(lr){8-9} \cmidrule(lr){10-11} \cmidrule(lr){12-13} \cmidrule(lr){14-15} \cmidrule(l){16-17}
 & EM $\uparrow$ & F1 $\uparrow$ & EM $\uparrow$ & F1 $\uparrow$ & EM $\uparrow$ & F1 $\uparrow$ & EM $\uparrow$ & F1 $\uparrow$ & EM $\uparrow$ & F1 $\uparrow$ & EM $\uparrow$ & F1 $\uparrow$ & EM $\uparrow$ & F1 $\uparrow$& EM$ \uparrow$ & F1 $\uparrow$\\
\midrule
REL & 3.778 & 4.684 & 6.666 & 7.470 & 5.999 & 6.628 & 5.111 & 6.571 & 2.889 & 3.656 & 3.999 & 3.424 & 7.555 & 7.519 & 6.444 & 6.451 \\
IRR & 2.445 & 4.812 & 6.000 & 6.562 & 6.222 & 7.427 & 2.889 & 4.671 & 2.000 & 4.476 & 5.778 & 5.895 & 4.889 & 6.466 & 5.778 & 6.866 \\
REL-1 & 2.667 & 4.459 & 8.444 & 9.465 & 3.333 & 4.018 & 4.222 & 4.786 & 5.334 & 6.104 & 5.555 & 6.261 & 5.778 & 5.515 & 1.445 & 0.943 \\
SPLIT & 4.222 & 5.248 & 6.222 & 7.045 & 7.112 & 6.315 & 6.445 & 6.063 & 11.111 & 11.244 & 10.000 & 9.460 & 7.556 & 5.700 & 5.111 & 5.182 \\
\bottomrule
\end{tabular}%
}
\caption{Client-specific performance gains (EM and F1) of CoRAG over RAG (Local) for various passage store configurations in the CRAB benchmark.}
\label{tab:appendix-client-incentives}
\end{table*}

\section{Formalizing Client Incentives}

\label{app:formal-incentives}

The collaborative nature of CoRAG introduces a novel tension between maximizing individual utility and contributing to the collective knowledge base. Unlike traditional collaborative learning, CoRAG requires clients to strategically decide which passages to contribute, balancing potential improvements from accessing a larger passage pool against the risk of incorporating hard negatives from other clients.

\paragraph{Definitions and Notation}
Let $N$ be the number of clients. For each client $i \in [N]$, we define:

\begin{itemize} [noitemsep, leftmargin=10pt, topsep=2pt, partopsep=2pt]
    \item $D_i$: The local training data of client $i$.
 \item $P_i$: The set of all passages available to client $i$.
\item $R_i$: The set of all passages relevant to client $i$'s training data $D_i$. Note that $R_i$ is not necessarily a subset of $P_i$.
\item $HN_i$: The set of all hard negative passages for client $i$. These are passages that appear relevant to client $i$'s retriever but do not contain the correct answer for $D_i$.
\item $IR_i$: The set of all irrelevant passages for client $i$, i.e., passages that are neither in $R_i$ nor in $HN_i$.
\end{itemize}

\noindent For any set of passages $P$ and client $i$, we define:
\begin{itemize} [noitemsep, leftmargin=10pt, topsep=2pt, partopsep=2pt]
    \item $R_i(P) = P \cap R_i$: The set of passages in $P$ that are relevant to client $i$.
\item $HN_i(P) = P \cap HN_i$: The set of hard negative passages in $P$ for client $i$.
\item $IR_i(P) = P \cap IR_i$: The set of irrelevant passages in $P$ for client $i$.
\end{itemize}

\paragraph{The CoRAG Participation Game}
We define the CoRAG participation game as follows:
\begin{definition}[The CoRAG Participation Game]
\label{def:corag_game}
The CoRAG participation game is a game with $N$ players (clients), where each player $i \in [N]$ chooses an action $a_i \in {0, 1}$: not contributing ($a_i=0$) or contributing ($a_i = 1$) their passage set $P_i$ to the shared store $P_{shared}$. Given an action profile $a = (a_1, \dots, a_N)$, player $i$'s payoff is defined as their utility:
\begin{equation}
\small
\label{eqn:utility}
U_i(a) = f_i(P_i \cup P_{shared}(a)) - f_i(P_i) - c_i a_i.
\end{equation}
Here, $f_i(P)$ denotes the performance of player $i$'s model when trained using passages $P$, $c_i>0$ represents the cost incurred by client $i$ for contributing, and $P_{shared}(a) = \bigcup_{j: a_j = 1} P_j$ is the shared store given the action profile $a$.
\end{definition}
We approximate the performance $f_i(P)$ as:
\begin{equation}
\small
\label{eqn:performance_approx}
f_i(P) \approx \alpha |R_i(P)| - \beta |HN_i(P)| + \gamma |IR_i(P)|,
\end{equation}
where coefficients $\alpha$, $\beta$, and $\gamma > 0$ capture the impact of each passage type on performance, with $\alpha > \gamma > \beta$.
\begin{definition}[Nash Equilibria in the CoRAG Game]
\label{def:corag_nash}
An action profile $a^*=(a^*_1,\dots, a^*_N)$ is a pure strategy Nash equilibrium of the CoRAG participation game if, for each player $i \in [N]$ and every action $a_i \in \{0,1\}$, $U_i(a^*_i, a^*_{-i}) \ge U_i(a_i, a^*_{-i})$.
\end{definition}

\paragraph{Analysis of Client Participation}
For a given action profile $a$, define:
\begin{itemize} [noitemsep, leftmargin=10pt, topsep=2pt, partopsep=2pt]
    \item $C(a) = \{ j \in [N] : a_j = 1\}$: The set of participating clients.
    \item $P_{shared}(a) = \bigcup_{j \in C(a)} P_j$: The shared store given action profile $a$.
\end{itemize}

A client $i$ participates in a Nash equilibrium $a^*$ if and only if:
\begin{equation}
\small
\label{eqn:cond_participate}
\begin{split}
U_i(1, a^*_{-i}) &\geq U_i(0, a^*_{-i}) \\
\iff f_i(P_i \cup P_{shared}(a^*)) &- f_i(P_i) \geq c_i
\end{split}
\end{equation}

Conversely, a client $i$ does not participate in a Nash equilibrium $a^*$ if and only if:
\begin{equation}
\small
\label{eqn:cond_not_participate}
\begin{split}
U_i(0, a^*_{-i}) &> U_i(1, a^*_{-i}) \\
\iff f_i(P_i \cup P_{shared}(a^*)) &- f_i(P_i) < c_i
\end{split}
\end{equation}

These conditions show that a client participates only if the performance gain from accessing the shared store exceeds their contribution cost. If the performance gain is less than the cost, the client will choose not to participate and will only use their local passages.

Using our performance approximation, we can expand the participation condition:
\begin{equation}
\small
\begin{split}
& \alpha |R_i(P_{shared}(a^*) \setminus P_i)| \\
& - \beta |HN_i(P_{shared}(a^*) \setminus P_i)| \\
& + \gamma |IR_i(P_{shared}(a^*) \setminus P_i)| \geq c_i
\end{split}
\end{equation}

The benefit of participation depends on the composition of the shared store relative to the client's local passages. Clients must weigh the potential gain from new relevant passages against the risk of incorporating hard negatives and the impact of irrelevant passages. Clients with many unique relevant passages may be less inclined to participate to maintain their competitive advantage. The equilibrium behavior of clients in this game depends on the distribution of passage types across clients and the individual participation costs.

\paragraph{Mechanisms for Encouraging Participation}
To address the tension between individual utility and contributing to the collective knowledge base, we propose the following mechanisms: \\
   \textbf{1. Contribution-Based Rewards:} We introduce a reward function that incentivizes clients to contribute high-quality passages:

    \begin{definition}[Reward Allocation Mechanism]
    \label{def:reward_mech}
    For a given action profile $a$, let $C(a) = \{j \in [N] : a_j = 1\}$ be the set of participating clients. The reward for client $i$ is:
    \begin{equation}
    \small
    r_i(a) =
    \begin{cases}
    \rho \cdot (|R_i \cap P_i| + \gamma |IR_i \cap P_i|) \cdot |C(a) \setminus \{i\}|, \\
    \quad \quad \quad \quad \quad \quad  \quad  \quad  \quad \quad  \quad \quad 
     \quad  \text{if } a_i=1 \\
    0, \quad \text{if } a_i=0
    \end{cases}
    \end{equation}
    where $\rho > 0$ is a scaling factor.
    \end{definition}

    This mechanism rewards participating clients based on the quality of their contributions (relevant and irrelevant passages) and the number of other participating clients. The inclusion of irrelevant passages in the reward calculation reflects their value in improving retrieval performance. \\

    \noindent \textbf{2. Tiered Access Levels:} We implement a tiered access system based on the quality and quantity of a client's contributions:

    \begin{equation}
    \small
    access_i = \min(1, \frac{|P_i|}{k \cdot \text{avg}_{j \in C(a)} |P_j|})
    \end{equation}

    where $k > 0$ is a parameter controlling the strictness of the access policy. This mechanism provides clients who contribute more passages with broader access to the shared store, incentivizing larger contributions. \\

    \noindent \textbf{3. Reputation Systems:} We establish a reputation system that tracks clients' contribution history:

    \begin{equation}
    \small
    reputation_i = \frac{|R_i \cap P_i| - \beta |HN_i \cap P_i|}{|P_i|}
    \end{equation}

    This reputation score balances the proportion of relevant passages a client contributes against the proportion of hard negatives, weighted by $\beta$ to reflect their relative impact on model performance.

\paragraph{CoRAG Game with Incentive Mechanisms}
Incorporating these mechanisms, we define a modified CoRAG game:

\begin{definition}[CoRAG Game with Incentive Mechanisms]
The modified CoRAG game with incentive mechanisms is defined as in Definition \ref{def:corag_game}, but with player $i$'s payoff defined as:
\begin{equation}
\small
\widetilde{U}_i(a) = U_i(a) + r_i(a) + v_i(access_i) + w_i(reputation_i),
\end{equation}
where $r_i(a)$ is the reward from Definition \ref{def:reward_mech}, $v_i(\cdot)$ and $w_i(\cdot)$ are non-decreasing functions representing the value player $i$ assigns to their access level and reputation, respectively.
\end{definition}

 The contribution-based reward encourages participation by compensating clients for the value they add to the shared store. Tiered access levels provide an additional incentive for clients to contribute more passages, while the reputation system introduces a long-term incentive for consistent, high-quality contributions.

This formalization provides a foundation for understanding the strategic considerations of clients in CoRAG and for designing effective incentive structures. Future work could focus on empirically evaluating these mechanisms and analyzing their impact on the Nash equilibria of the modified game.

\end{document}